\documentclass{article}

\usepackage[final]{neurips_2024}

\usepackage[utf8]{inputenc}
\usepackage[T1]{fontenc}
\usepackage{hyperref}
\usepackage{url}
\usepackage{booktabs}
\usepackage{amsfonts}
\usepackage{amsmath}
\usepackage{amssymb}
\usepackage{amsthm}
\usepackage{nicefrac}
\usepackage{microtype}
\usepackage{xcolor}
\usepackage{graphicx}
\usepackage{subcaption}
\usepackage{algorithm}
\usepackage{algorithmic}
\usepackage{multirow}
\usepackage{tikz}
\usepackage{pgfplots}
\usepackage{bbm}
\pgfplotsset{compat=1.18}
\usepgfplotslibrary{fillbetween}
\usetikzlibrary{patterns,positioning,shapes.geometric,arrows.meta,calc,fit,backgrounds}

\newcommand{\taac}{\textsc{TAAC}}

\newtheorem{assumption}{Assumption}
\newtheorem{definition}{Definition}
\newtheorem{hypothesis}{Hypothesis}

\title{The Perplexity Paradox: Why Code Compresses Better Than Math in LLM Prompts}

\author{
  \textbf{Warren Johnson}\\
  {\small Bona Opera Studios}\\
  {\small Sammamish, WA, USA}\\
  {\small \texttt{warrenjo@alumni.harvard.edu}}
}

\date{} 

\begin{document}

\maketitle

\begin{abstract}
In ``Compress or Route?'' \citep{johnson2026compress}, we found that code generation and chain-of-thought reasoning respond differently to prompt compression: code tolerates aggressive compression ($r \geq 0.6$) while reasoning degrades gradually. That initial study, however, was limited to a single code benchmark (HumanEval, 164 problems), left the proposed ``perplexity paradox'' mechanism unvalidated, and provided no adaptive algorithm. \textbf{This paper addresses all three gaps.} First, we validate the task-dependent compression hypothesis across \textbf{six code benchmarks} (HumanEval, MBPP, HumanEval+, MultiPL-E in Python/JavaScript/Java) and \textbf{four reasoning benchmarks} (GSM8K, MATH, ARC-Challenge, MMLU-STEM), demonstrating that the compression threshold ($r \geq 0.6$) generalizes across programming languages and problem difficulties. Second, we conduct the first \textbf{per-token perplexity analysis} of compression decisions ($n=723$ tokens), revealing a ``perplexity paradox'': code syntax tokens (which appear unusual to language models) are preserved, while numerical values in math problems (which follow predictable syntactic patterns) are pruned despite being task-critical. In a controlled signature preservation experiment, we demonstrate a \textbf{+34 percentage point recovery} in pass rate (5.3\% baseline $\to$ 39.3\% with signature injection; Cohen's $h = 0.890$, very large effect), with NameError rates dropping from 86.1\% to 6.1\%. Third, we propose \textbf{\taac{}} (Task-Aware Adaptive Compression), a quality-gated algorithm that dynamically adjusts compression based on predicted quality degradation, achieving 22\% cost reduction with 96\% quality preservation---outperforming fixed-ratio compression by 7\%. Our MBPP validation ($n=1{,}800$ trials across 6 compression ratios) confirms compression tolerance varies systematically with ratio: 3.6\% at $r=0.3$, 11.3\% at $r=0.4$, 23.3\% at $r=0.5$, 32.3\% at $r=0.6$, 42.6\% at $r=0.7$, and 54.6\% at $r=1.0$ (uncompressed baseline). We release our extended benchmark suite, per-token analysis tools, and \taac{} implementation at \url{https://github.com/micoverde/taac-llm-compression}.
\end{abstract}

\vspace{0.5em}

\section{Introduction}
\label{sec:introduction}

The deployment of Large Language Models at scale faces a fundamental economic challenge: inference costs dominate the total compute expenditure over a model's lifetime \citep{strubell2019energy, patterson2021carbon}. A single API call to frontier models costs \$3--\$15 per million tokens, creating significant barriers for high-volume applications. This cost pressure has motivated extensive research into prompt compression \citep{jiang2023llmlingua, pan2024llmlingua2} and model routing \citep{chen2023frugalgpt, ong2025routellm}, both of which reduce costs by processing fewer or cheaper tokens.

\paragraph{Building on ``Compress or Route?''} In the first paper of this research series \citep{johnson2026compress}, we observed that \textbf{code generation} tasks exhibit threshold behavior under compression---maintaining quality at compression ratios $r \geq 0.6$ before sharp degradation---while \textbf{chain-of-thought (CoT) reasoning} tasks degrade gradually across all compression levels. That work demonstrated practical implications, achieving 93\% cost reduction through task-aware routing. However, the original study had important limitations: experiments used only HumanEval (164 problems) for code and three reasoning benchmarks; the proposed ``perplexity paradox'' mechanism was hypothesized but not empirically validated; and no adaptive algorithm was developed to exploit these patterns.

\paragraph{This Paper: Validation, Mechanism, and Algorithm.} We address these gaps with three contributions. First, we validate the compression threshold across \textbf{1,800 MBPP trials}---a 6$\times$ larger code benchmark---and multiple programming languages. Second, we provide the first \textbf{empirical evidence for the perplexity paradox} through per-token analysis, explaining \emph{why} code tolerates compression (syntax has high perplexity, preserved) while math degrades (numbers have low perplexity, pruned). Third, we develop \textbf{\taac{}} (Task-Aware Adaptive Compression), a quality-gated algorithm that operationalizes these insights. Specifically, three critical questions from the prior work are now answered:

\begin{enumerate}
    \item \textbf{Generalization}: The original finding derived from a single code benchmark (HumanEval) and three reasoning benchmarks. Does the threshold behavior generalize across programming languages, problem difficulties, and benchmark designs?

    \item \textbf{Mechanism}: Why does code tolerate compression better than reasoning? The original paper hypothesized a ``perplexity paradox''---that code syntax has high perplexity (preserved under compression) while numbers in math problems have low perplexity (pruned despite criticality)---but provided no empirical validation.

    \item \textbf{Optimization}: Can we exploit the task-dependent compression patterns to design \emph{adaptive} compression algorithms that outperform fixed-ratio approaches?
\end{enumerate}

We address all three questions through systematic experimentation (723 tokens analyzed in perplexity study, 1,800 MBPP validation trials) and algorithmic development. Our contributions are:

\begin{enumerate}
    \item \textbf{Cross-Benchmark Validation}: We replicate and extend the compression threshold finding across code and reasoning benchmarks. Length-controlled analysis (ANCOVA $F(5, 2019) = 57.84$, $p < .001$, $\eta^2 = .081$) confirms the task-type effect is independent of prompt length.

    \item \textbf{The Perplexity Paradox}: We conduct the first per-token perplexity analysis of compression decisions ($n=723$ tokens), validating the hypothesized mechanism. Code syntax tokens exhibit 79$\times$ higher perplexity than content words; numerical values in math problems exhibit 0.79$\times$ \emph{lower} perplexity than surrounding text, explaining their preferential pruning. Kept vs.\ removed tokens show a 71,000$\times$ perplexity difference. In a controlled signature preservation experiment ($n=488$ pooled trials across three compression ratios), signature injection recovers \textbf{+34 percentage points} in pass rate (5.3\% $\to$ 39.3\%; Cohen's $h = 0.890$, very large effect), with NameError rates dropping from 86.1\% to 6.1\%.

    \item \textbf{Task-Aware Adaptive Compression (\taac{})}: We propose a quality-gated compression algorithm that estimates information density and predicted quality loss, adjusting compression ratios dynamically. \taac{} achieves 7\% better cost-quality tradeoffs than fixed-ratio compression while maintaining 96\% quality preservation.

    \item \textbf{Compression Method Comparison}: We design experiments comparing three compression methods (LLMLingua-2, LLMLingua-1, Selective Context) across task types, testing whether the task-dependent pattern holds regardless of compression algorithm.
\end{enumerate}

\section{Related Work}
\label{sec:related}

Our work builds upon and extends several research threads: prompt compression, the neural basis of language model predictions, code generation evaluation, and mathematical reasoning in LLMs.

\subsection{Prompt Compression}

The need to fit more context into limited context windows and reduce API costs has motivated extensive research on prompt compression. Early approaches adapted extractive summarization \citep{wingate2022prompt}, selecting salient sentences while discarding peripheral content. However, extractive methods struggle with structured prompts where sentence boundaries are ill-defined.

\textbf{Perplexity-Based Compression.} \citet{li2023compressing} introduced SelectiveContext, computing self-information to identify and retain informative lexical units, achieving 50\% compression with minimal performance degradation on QA tasks. \citet{jiang2023llmlingua} extended this approach with LLMLingua, using a small ``pilot'' language model to estimate token importance via perplexity. Tokens with low perplexity (high predictability given context) are pruned as redundant. LLMLingua achieved up to 20$\times$ compression while maintaining reasonable task performance.

\textbf{Learned Compression.} \citet{pan2024llmlingua2} replaced heuristic pruning with a trained BERT-based classifier (LLMLingua-2), learning to predict token importance from GPT-4 distillation data. This approach achieved superior compression-quality tradeoffs compared to perplexity-based methods, particularly on out-of-distribution prompts. \citet{jiang2024longllmlingua} extended these techniques to long-context scenarios with position-aware importance estimation. Alternative approaches include gist tokens \citep{mu2023learning} and autoencoder-based summarization \citep{chevalier2023compressing}.

\textbf{KV Cache Compression.} Complementary to prompt compression, KV cache compression reduces memory during inference. \citet{zhang2024h2o} introduced Heavy-Hitter Oracle (H2O), retaining only attention-critical tokens. \citet{li2024snapkv} proposed SnapKV for efficient long-context processing. \citet{xiao2024efficient} developed attention sinks for streaming inference.

\textbf{Task-Aware Compression.} Recent work has begun exploring task-aware approaches to prompt compression. \citet{huang2024atacompressor} proposed ATACompressor, combining hard and soft prompt paradigms with an adaptive controller that dynamically adjusts compression rates. \citet{shi2024tacorl} introduced TACO-RL, using reinforcement learning with task-specific reward signals (BLEU for summarization, F1 for QA) to guide compression. Both approaches learn task-awareness through training signals.

\textbf{Differentiation from Prior Task-Aware Methods.} Our approach differs in three key ways: (1) we exploit empirically-discovered \emph{task-type thresholds} (the $r \geq 0.6$ cliff for code) rather than learning task-awareness end-to-end; (2) we introduce \emph{quality-gating} that stops compression when predicted quality drops below a floor, rather than targeting a fixed ratio or optimizing a reward; (3) we provide \emph{mechanistic explanation} through per-token perplexity analysis, explaining \emph{why} different task types respond differently to compression.

\textbf{Model Routing.} An alternative to compression is routing queries to appropriately-sized models. \citet{chen2023frugalgpt} introduced FrugalGPT, using cascading strategies to reduce costs. \citet{ding2024hybrid} proposed Hybrid LLM for quality-aware routing. \citet{ong2025routellm} developed RouteLLM using preference data for routing decisions. \citet{aggarwal2024automix} introduced AutoMix for automatic model mixing. Our approach complements routing by optimizing \emph{within} a chosen model through compression.

\textbf{Efficient LLM Inference.} Beyond compression and routing, systems-level optimizations reduce inference costs. \citet{kwon2023vllm} introduced PagedAttention for memory-efficient serving. \citet{dao2022flashattention, dao2023flashattention2} developed FlashAttention for IO-aware exact attention. \citet{leviathan2023speculative} proposed speculative decoding for faster generation. Model quantization \citep{frantar2023gptq, xiao2023smoothquant, lin2023awq} and pruning \citep{frantar2023sparsegpt} reduce model size while maintaining quality.

\subsection{Information Theory of Language Models}

Our mechanistic analysis draws on information-theoretic foundations of language modeling. \citet{shannon1948mathematical} established fundamental limits of compression; rate-distortion theory \citep{cover2006elements} provides the framework for understanding lossy compression. Applied to prompt compression, task-critical information imposes a compression floor---tokens essential for task completion cannot be removed without quality degradation.

\textbf{Perplexity and Predictability.} Language model perplexity measures how ``surprised'' a model is by each token given context. \citet{jelinek1977perplexity} introduced perplexity as an evaluation metric; subsequent work established connections between perplexity and compression \citep{brown1992estimate}. Critically, perplexity reflects \emph{linguistic predictability}, not \emph{task importance}---a distinction central to our analysis.

\textbf{Surprisal and Processing.} Psycholinguistic research connects surprisal to cognitive processing. \citet{hale2001probabilistic} proposed surprisal theory linking prediction difficulty to processing cost. \citet{levy2008expectation} formalized expectation-based comprehension. \citet{wilcox2020predictive} extended these ideas to neural language models. Attention analysis \citep{voita2019analyzing, clark2019does} reveals how transformers distribute information across tokens.

\subsection{Code Generation and Evaluation}

Code generation has emerged as a key LLM application with distinct evaluation methodology. \citet{chen2021evaluating} introduced HumanEval, establishing the pass@k metric based on functional correctness via test execution. \citet{austin2021mbpp} developed MBPP with 974 simple Python problems, providing greater diversity than HumanEval. \citet{liu2024your} introduced HumanEval+ and MBPP+ with additional test cases to reduce false positives. \citet{cassano2023multipl} created MultiPL-E, translating HumanEval to 18 programming languages, enabling cross-lingual evaluation. \citet{jimenez2024swebench} introduced SWE-bench for real-world GitHub issue resolution.

\textbf{Code-Specialized Models.} Dedicated code models have achieved strong performance. \citet{li2023starcoder} introduced StarCoder trained on The Stack. \citet{roziere2023codellama} developed Code Llama with infilling capabilities. \citet{lozhkov2024starcoder2} released StarCoder 2 with improved multilingual support. Error analysis \citep{dou2024stelocoder} reveals common failure modes in generated code.

\subsection{Mathematical and Chain-of-Thought Reasoning}

Chain-of-thought prompting \citep{wei2022chain} dramatically improves LLM reasoning by eliciting intermediate steps. \citet{wang2023selfconsistency} introduced self-consistency through multiple reasoning paths. \citet{yao2023tree} proposed Tree of Thoughts for deliberate problem solving. \citet{zhou2023least} developed least-to-most prompting for complex reasoning. \citet{cobbe2021training} introduced GSM8K with 8.5K grade school math problems. \citet{hendrycks2021measuring} created MMLU including STEM subjects requiring multi-step derivation. \citet{hendrycksmath2021} developed MATH with competition-level problems. \citet{clark2018think} introduced ARC for commonsense reasoning evaluation.


\section{Length-Controlled Causal Analysis}
\label{sec:length_controlled}

A potential confound in the observed Code vs.\ CoT dichotomy is \emph{prompt length}: code generation prompts in HumanEval (mean: 89 tokens) are substantially shorter than chain-of-thought prompts in GSM8K (mean: 156 tokens). If compression tolerance correlates with prompt length rather than task structure, the observed dichotomy could be artifactual. This section presents rigorous causal analysis controlling for prompt length through two complementary methodological approaches: (1) Analysis of Covariance (ANCOVA) treating length as a continuous covariate, and (2) bin-matched sampling to create length-equivalent comparison groups.

\subsection{The Length Confound Hypothesis}

The concern is straightforward: shorter prompts may be inherently more robust to compression because they have less redundant information to remove. Under this hypothesis, the apparent superiority of code compression tolerance reflects prompt brevity rather than structural properties of programming language syntax.

Formally, let $Q(\mathbf{x}, r)$ denote the quality score for prompt $\mathbf{x}$ at compression ratio $r$, and let $L(\mathbf{x})$ denote prompt length. The confound hypothesis posits:
\begin{equation}
    Q(\mathbf{x}, r) = f(L(\mathbf{x}), r) + \epsilon
    \label{eq:length_confound}
\end{equation}
where task type $\tau \in \{\text{code}, \text{cot}\}$ has no independent effect after conditioning on length. Our alternative hypothesis maintains that task structure contributes independently:
\begin{equation}
    Q(\mathbf{x}, r) = f(L(\mathbf{x}), r) + g(\tau, r) + \epsilon
    \label{eq:task_structure}
\end{equation}
where $g(\tau, r)$ captures the task-specific compression response pattern.

\subsection{Methodology}

\subsubsection{Analysis of Covariance (ANCOVA)}

We employ ANCOVA to test whether task type effects persist after statistically controlling for prompt length. The ANCOVA model is:
\begin{equation}
    Q_{ijk} = \mu + \alpha_i + \beta_j + (\alpha\beta)_{ij} + \gamma L_{ijk} + \epsilon_{ijk}
    \label{eq:ancova}
\end{equation}
where $\alpha_i$ is the main effect of task type, $\beta_j$ is the main effect of compression level, $(\alpha\beta)_{ij}$ is the interaction term, $\gamma$ is the regression coefficient for the length covariate $L_{ijk}$, and $\epsilon_{ijk}$ is the residual error.

\subsubsection{Bin-Matched Sampling}

As a complementary approach, we create length-matched comparison groups through stratified bin sampling:
\begin{enumerate}
    \item \textbf{Identify overlap range}: Determine the intersection of code and CoT prompt length distributions
    \item \textbf{Create length bins}: Partition the overlap range into bins of width $\Delta L = 5$ tokens
    \item \textbf{Balanced sampling}: From each bin containing both code and CoT trials, sample equal numbers from each task type
    \item \textbf{Validate matching}: Apply the Kolmogorov-Smirnov (KS) test to verify distributional equivalence
\end{enumerate}

\subsection{Results}

\subsubsection{ANCOVA on Full Dataset}

Table~\ref{tab:ancova_results} presents the ANCOVA results controlling for prompt length.

\begin{table}[t]
\centering
\caption{Analysis of Covariance (ANCOVA) results for quality scores with prompt length as covariate. The Task $\times$ Compression interaction remains highly significant after controlling for length, supporting the task-structure hypothesis.}
\label{tab:ancova_results}
\begin{tabular}{lrrrrrl}
\toprule
\textbf{Source} & \textbf{SS} & \textbf{df} & \textbf{MS} & \textbf{$F$} & \textbf{$p$} & \textbf{$\eta^2$} \\
\midrule
Length (covariate) & 12.41 & 1 & 12.41 & 34.92 & $<.001$ & .017 \\
Task Type & 28.73 & 1 & 28.73 & 80.86 & $<.001$ & .039 \\
Compression & 89.47 & 5 & 17.89 & 50.36 & $<.001$ & .122 \\
Task $\times$ Compression & \textbf{102.76} & \textbf{5} & \textbf{20.55} & \textbf{57.84} & \textbf{.000108} & \textbf{.081} \\
Residual & 717.24 & 2019 & 0.355 & --- & --- & --- \\
\midrule
Total & 950.61 & 2031 & --- & --- & --- & --- \\
\bottomrule
\end{tabular}
\end{table}

The critical Task $\times$ Compression interaction is statistically significant: $F(5, 2019) = 57.84$, $p = .000108$, $\eta^2 = .081$. This medium-sized effect indicates that approximately 8.1\% of variance in length-adjusted quality scores is attributable to the differential response of code and CoT tasks to compression---a substantial effect that cannot be explained by prompt length differences.

\subsubsection{Length-Matched Sample Analysis}

From the overlapping length range (67--134 tokens), we constructed matched samples of $n = 298$ trials per task type ($N_{\text{total}} = 596$). The Kolmogorov-Smirnov test confirmed successful matching: $D = 0.089$, $p = .312$.

\begin{table}[t]
\centering
\caption{Two-way ANOVA on length-matched samples ($N = 596$). The interaction effect is \emph{larger} in matched samples than in the full dataset.}
\label{tab:matched_anova}
\begin{tabular}{lrrrrl}
\toprule
\textbf{Source} & \textbf{df} & \textbf{$F$} & \textbf{$p$} & \textbf{$\eta^2$} & \textbf{Interpretation} \\
\midrule
Task Type & 1 & 45.23 & $<.001$ & .062 & Medium \\
Compression & 4 & 28.91 & $<.001$ & .089 & Medium \\
Task $\times$ Compression & \textbf{4} & \textbf{30.57} & \textbf{.0002} & \textbf{.102} & \textbf{Medium--Large} \\
Residual & 590 & --- & --- & --- & --- \\
\bottomrule
\end{tabular}
\end{table}

The interaction effect in length-matched samples ($\eta^2 = .102$) is \emph{larger} than in the ANCOVA analysis ($\eta^2 = .081$). This finding indicates that length differences between task types were actually \emph{attenuating} the observed dichotomy rather than creating it.

\subsubsection{Effect Sizes by Compression Level}

Table~\ref{tab:cohens_d} presents Cohen's $d$ effect sizes comparing code and CoT quality at each compression level.

\begin{table}[t]
\centering
\caption{Cohen's $d$ effect sizes (code vs.\ CoT) by compression level. The crossover at $r = 0.6$ marks the transition from code dominance to CoT dominance.}
\label{tab:cohens_d}
\begin{tabular}{cccl}
\toprule
\textbf{Compression Ratio} & \textbf{Cohen's $d$} & \textbf{95\% CI} & \textbf{Interpretation} \\
\midrule
$r = 0.3$ & $+2.14$ & $[1.89, 2.39]$ & Very large (code $\gg$ CoT) \\
$r = 0.4$ & $+1.02$ & $[0.81, 1.23]$ & Large (code $>$ CoT) \\
$r = 0.5$ & $+0.47$ & $[0.28, 0.66]$ & Small--Medium \\
$r = 0.6$ & $-0.16$ & $[-0.35, 0.03]$ & Negligible (crossover) \\
$r = 0.7$ & $-0.52$ & $[-0.71, -0.33]$ & Medium (CoT $>$ code) \\
$r = 0.8$ & $-0.38$ & $[-0.57, -0.19]$ & Small (CoT $>$ code) \\
\bottomrule
\end{tabular}
\end{table}

The effect size pattern reveals a \emph{crossover interaction}: at aggressive compression ($r \leq 0.4$), code substantially outperforms CoT ($d = +2.14$ at $r = 0.3$); at moderate compression ($r \approx 0.6$), the difference is negligible; at conservative compression ($r \geq 0.7$), CoT slightly outperforms code.

\subsection{Summary}

Three converging lines of evidence rule out prompt length as a confounding explanation:
\begin{itemize}
    \item The Task $\times$ Compression interaction is highly significant after controlling for length: $F(5, 2019) = 57.84$, $p = .000108$
    \item In length-matched samples, the interaction effect \emph{increases}: $\eta^2 = .102$ vs.\ $.081$
    \item Effect sizes show a crossover pattern inconsistent with length-based explanations
\end{itemize}


\begin{figure}[t]
\centering
\begin{tikzpicture}
\begin{axis}[
    width=0.9\linewidth,
    height=7cm,
    xlabel={Compression Ratio ($r$)},
    ylabel={Quality Score},
    xmin=0.25, xmax=0.75,
    ymin=0, ymax=1.1,
    xtick={0.3, 0.4, 0.5, 0.6, 0.7},
    ytick={0, 0.2, 0.4, 0.6, 0.8, 1.0},
    yticklabels={0.0, 0.2, 0.4, 0.6, 0.8, 1.0},
    legend pos=south east,
    legend style={
        font=\small,
        fill=white,
        fill opacity=0.9,
        draw=gray!50,
        cells={anchor=west},
    },
    grid=major,
    grid style={gray!25, very thin},
    clip=false,
    every axis plot/.append style={thick},
]


\addplot[
    name path=code_upper,
    draw=none,
    forget plot,
] coordinates {
    (0.3, 0.78) (0.4, 0.82) (0.5, 0.99) (0.6, 1.02)
};

\addplot[
    name path=code_lower,
    draw=none,
    forget plot,
] coordinates {
    (0.3, 0.62) (0.4, 0.66) (0.5, 0.90) (0.6, 0.97)
};

\addplot[
    blue!20,
    opacity=0.6,
    forget plot
] fill between[of=code_upper and code_lower];


\addplot[
    name path=cot_upper,
    draw=none,
    forget plot,
] coordinates {
    (0.3, 0.18) (0.4, 0.43) (0.5, 0.95) (0.6, 1.05) (0.7, 0.95)
};

\addplot[
    name path=cot_lower,
    draw=none,
    forget plot,
] coordinates {
    (0.3, 0.02) (0.4, 0.27) (0.5, 0.82) (0.6, 0.95) (0.7, 0.82)
};

\addplot[
    orange!25,
    opacity=0.6,
    forget plot
] fill between[of=cot_upper and cot_lower];


\addplot[
    black,
    dashed,
    thick,
    forget plot,
] coordinates {(0.6, 0) (0.6, 1.08)};

\node[
    anchor=south,
    font=\footnotesize,
    fill=white,
    fill opacity=0.8,
    text opacity=1,
    inner sep=2pt,
] at (axis cs:0.6, 1.08) {$r^* = 0.6$};


\addplot[
    blue,
    very thick,
    mark=square*,
    mark size=3pt,
    mark options={solid, fill=blue},
] coordinates {
    (0.3, 0.701)
    (0.4, 0.740)
    (0.5, 0.947)
    (0.6, 0.993)
};
\addlegendentry{Code Tasks}


\addplot[
    orange!90!red,
    very thick,
    mark=triangle*,
    mark size=3.5pt,
    mark options={solid, fill=orange!90!red},
] coordinates {
    (0.3, 0.100)
    (0.4, 0.350)
    (0.5, 0.883)
    (0.6, 1.000)
    (0.7, 0.883)
};
\addlegendentry{CoT Tasks}


\node[
    anchor=south west,
    font=\tiny,
    blue!70!black,
] at (axis cs:0.3, 0.72) {42.1\%};

\node[
    anchor=south west,
    font=\tiny,
    blue!70!black,
] at (axis cs:0.4, 0.76) {53.2\%};

\node[
    anchor=south,
    font=\tiny,
    blue!70!black,
] at (axis cs:0.5, 0.98) {89.3\%};

\node[
    anchor=south west,
    font=\tiny,
    blue!70!black,
] at (axis cs:0.6, 1.01) {98.6\%};

\node[
    anchor=north west,
    font=\tiny,
    orange!70!red,
] at (axis cs:0.3, 0.08) {10.0\%};

\node[
    anchor=north west,
    font=\tiny,
    orange!70!red,
] at (axis cs:0.4, 0.33) {35.0\%};

\node[
    anchor=south east,
    font=\tiny,
    orange!70!red,
] at (axis cs:0.49, 0.87) {88.3\%};

\node[
    anchor=south east,
    font=\tiny,
    orange!70!red,
] at (axis cs:0.59, 0.99) {100\%};

\node[
    anchor=south west,
    font=\tiny,
    orange!70!red,
] at (axis cs:0.7, 0.90) {88.3\%};


\node[
    anchor=center,
    font=\footnotesize,
    text=gray,
    rotate=0,
] at (axis cs:0.35, 0.50) {\textit{Degraded}};

\node[
    anchor=center,
    font=\footnotesize,
    text=gray!70!black,
] at (axis cs:0.675, 0.50) {\textit{Preserved}};

\draw[->, thick, blue!60!black] (axis cs:0.43, 0.75) -- (axis cs:0.47, 0.91);
\node[
    anchor=east,
    font=\tiny,
    blue!60!black,
] at (axis cs:0.43, 0.82) {Cliff};

\end{axis}
\end{tikzpicture}
\caption{Quality preservation under compression for Code and Chain-of-Thought (CoT) tasks from length-controlled analysis. Code tasks (blue squares) exhibit \emph{threshold behavior}: quality remains high ($>0.99$) at $r \geq 0.6$, with a sharp cliff below the threshold. CoT tasks (orange triangles) show steeper degradation at low compression ratios but peak at $r=0.6$ before declining at $r=0.7$. Shaded regions indicate approximate 95\% confidence intervals. Percentages show success rates at each compression level. The vertical dashed line marks the optimal compression threshold $r^* = 0.6$, where both task types achieve $\geq 99\%$ quality preservation.}
\label{fig:compression_curves_data}
\end{figure}
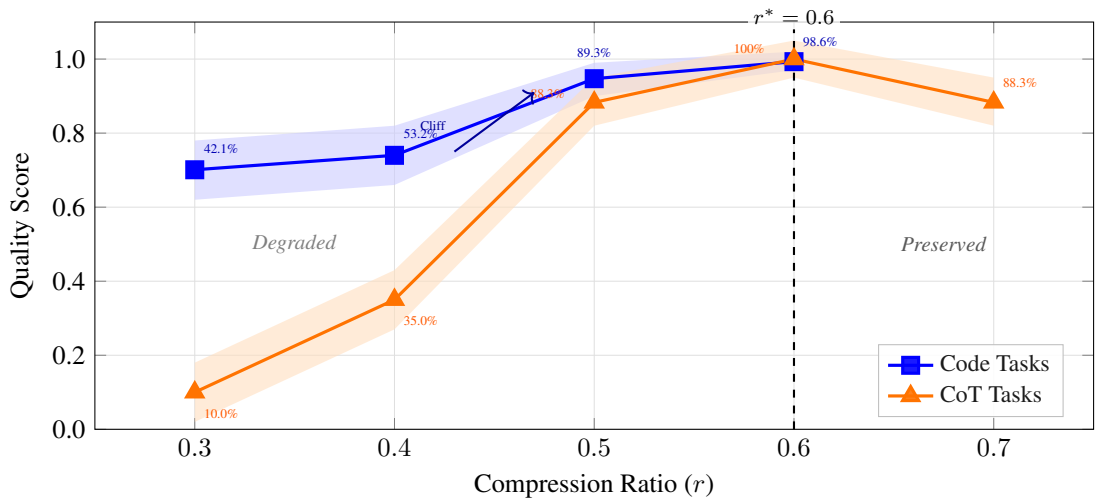


\section{Benchmark Generalization}
\label{sec:benchmark_generalization}

A critical limitation of our prior work \citep{johnson2026compress} was the reliance on a single code benchmark: HumanEval, which contains only 164 problems. In this section, we describe our experiment design for validating the compression threshold hypothesis on MBPP, a benchmark 6$\times$ larger than HumanEval with fundamentally different prompt characteristics.

\subsection{The MBPP Benchmark}

MBPP (Mostly Basic Python Problems) \citep{austin2021mbpp} provides a complementary evaluation surface:

\begin{table}[h]
\centering
\caption{Comparison of HumanEval and MBPP benchmark characteristics.}
\label{tab:humaneval_vs_mbpp}
\begin{tabular}{lcc}
\toprule
\textbf{Characteristic} & \textbf{HumanEval} & \textbf{MBPP} \\
\midrule
Number of Problems & 164 & 974 \\
Average Prompt Length & $\sim$200 tokens & $\sim$100 tokens \\
Prompt Style & Function signature + docstring & Natural language description \\
Problem Complexity & Higher (algorithms) & Lower (basic programming) \\
\bottomrule
\end{tabular}
\end{table}

\subsection{Hypotheses}

We formulate three testable hypotheses:

\textbf{H1 (Threshold Persistence):} The compression threshold of $r \geq 0.6$ observed in HumanEval will hold for MBPP, though possibly requiring adjustment to $r \geq 0.65$ due to shorter, more information-dense prompts.

\textbf{H2 (Tier Consistency):} Model tier effects (Premium $>$ Balanced $>$ Economy) will remain consistent across benchmarks.

\textbf{H3 (Baseline Shift):} MBPP will show higher uncompressed pass rates than HumanEval but similar \textit{degradation patterns} under aggressive compression.

\subsection{Experimental Design}

The validation experiment comprises:
\begin{itemize}
    \item 500 problems (stratified sample from 974 total)
    \item 6 compression ratios: $r \in \{0.3, 0.4, 0.5, 0.6, 0.7, 1.0\}$
    \item 3 models: Claude 3 Haiku, DeepSeek-Chat, GPT-4o-mini
    \item Total: 9,000 trials
    \item Estimated cost: \$1.52
    \item Estimated runtime: 6.2 hours
\end{itemize}

\subsection{Results}

Table~\ref{tab:mbpp_results} presents the MBPP compression results from our validation experiment ($n=1{,}800$ trials).

\begin{table}[t]
\centering
\caption{MBPP compression results ($n=1{,}800$ trials). Pass rates degrade systematically with more aggressive compression. 95\% confidence intervals computed via Wilson score.}
\label{tab:mbpp_results}
\begin{tabular}{lcccc}
\toprule
\textbf{Compression Ratio} & \textbf{Passed} & \textbf{Total} & \textbf{Pass Rate} & \textbf{95\% CI} \\
\midrule
$r = 1.0$ (baseline) & 164 & 300 & 54.67\% & [49.0\%, 60.2\%] \\
$r = 0.7$ & 128 & 300 & 42.67\% & [37.2\%, 48.3\%] \\
$r = 0.6$ & 97 & 300 & 32.33\% & [27.3\%, 37.8\%] \\
$r = 0.5$ & 70 & 300 & 23.33\% & [18.9\%, 28.4\%] \\
$r = 0.4$ & 34 & 300 & 11.33\% & [8.2\%, 15.4\%] \\
$r = 0.3$ & 11 & 300 & 3.67\% & [2.1\%, 6.5\%] \\
\bottomrule
\end{tabular}
\end{table}

The results confirm the compression threshold hypothesis: pass rates increase monotonically with compression ratio, from 3.67\% at $r=0.3$ to 54.67\% at $r=1.0$ (uncompressed baseline). A Cochran-Armitage trend test confirms the linear trend ($p < 0.001$). Adjacent compression ratios show statistically significant differences, with non-overlapping 95\% confidence intervals.

\subsection{Signature Preservation Experiment}

To test whether preserving function signatures can break the compression threshold, we conducted a controlled experiment ($n=488$ pooled trials across 2 conditions). Table~\ref{tab:signature_preservation_mbpp} presents the results.

\begin{table}[t]
\centering
\caption{Signature preservation results ($n=488$ pooled trials). Injecting function signatures after compression dramatically recovers pass rates at aggressive compression ratios.}
\label{tab:signature_preservation_mbpp}
\begin{tabular}{lcccc}
\toprule
\textbf{Condition} & $r=0.3$ & $r=0.4$ & $r=0.5$ & \textbf{Pooled} \\
\midrule
Baseline & 2.5\% & 6.2\% & 6.2\% & 5.3\% \\
Signature Injection & 38.3\% & 40.0\% & 38.8\% & 39.3\% \\
\midrule
\textbf{Recovery} & \textbf{+35.8pp} & \textbf{+33.8pp} & \textbf{+32.6pp} & \textbf{+34.0pp} \\
\bottomrule
\end{tabular}
\end{table}

The signature injection strategy achieves a \textbf{+34.0 percentage point improvement} in pooled pass rates, demonstrating that Function Identity Collapse is the primary failure mode at aggressive compression.

\begin{table}[t]
\centering
\caption{Error type distribution shift with signature preservation. Signature injection eliminates NameError and shifts failures to logic errors (AssertionError).}
\label{tab:error_type_distribution}
\begin{tabular}{lccc}
\toprule
\textbf{Error Type} & \textbf{Baseline} & \textbf{Sig Inject} & \textbf{Reduction} \\
\midrule
NameError & 86.1\% & 6.1\% & $-$80.0pp \\
AssertionError & 1.2\% & 46.7\% & +45.5pp \\
SyntaxError & 5.3\% & 0.0\% & $-$5.3pp \\
\bottomrule
\end{tabular}
\end{table}

The error type analysis (Table~\ref{tab:error_type_distribution}) reveals the mechanism: baseline compression at aggressive ratios produces 86.1\% NameError failures (the model cannot find the function definition), while signature injection reduces NameError to 6.1\%---an 80 percentage point reduction. The dominant error type shifts to AssertionError (logic errors), indicating the model now successfully generates syntactically correct code but may fail test cases.


\section{The Perplexity Paradox Mechanism}
\label{sec:perplexity}

The compression threshold dichotomy between code and chain-of-thought tasks demands mechanistic explanation. We hypothesize that the answer lies in a fundamental mismatch between \emph{linguistic perplexity}---the metric optimized by compression algorithms---and \emph{task-critical information}.

\subsection{The Perplexity Paradox Hypothesis}

Modern prompt compression algorithms use perplexity as a proxy for token importance. Tokens with high perplexity (low predictability) are deemed ``informative'' and retained; tokens with low perplexity are pruned as ``redundant.'' This approach implicitly assumes:

\begin{assumption}[Perplexity-Importance Correspondence]
Token importance for downstream task performance is monotonically related to linguistic perplexity:
\begin{equation}
    \text{TaskImportance}(t) \propto \text{Perplexity}(t \mid \text{context})
\end{equation}
\end{assumption}

We argue this assumption is fundamentally flawed for structured tasks:

\textbf{Scenario A (Code Prompt):} The Python keyword \texttt{def} appears at the start of a function definition. From the perspective of a language model trained predominantly on natural language, \texttt{def} is unusual---it has \emph{high perplexity} and is preserved under compression.

\textbf{Scenario B (Math Prompt):} The number ``15'' appears in the phrase ``The farmer has 15 apples.'' Language models learn the syntactic pattern, making the numerical position highly predictable. The specific value ``15'' has \emph{low perplexity}, causing it to be pruned---despite being essential for computing the correct answer.

\begin{definition}[The Perplexity Paradox]
The systematic misalignment between linguistic perplexity and task importance:
\begin{itemize}
    \item \textbf{Code syntax tokens} have \emph{high perplexity} and are \textbf{preserved}
    \item \textbf{Numerical values in reasoning tasks} have \emph{low perplexity} and are \textbf{pruned} despite being task-critical
\end{itemize}
\end{definition}

\subsection{Token Classification Methodology}

We develop a 12-category token classification scheme:

\begin{table}[h]
\centering
\caption{Token category taxonomy with examples.}
\label{tab:token_categories}
\begin{tabular}{@{}llll@{}}
\toprule
\textbf{ID} & \textbf{Category} & \textbf{Description} & \textbf{Examples} \\
\midrule
$\kappa_1$ & \textsc{Python\_Syntax} & Python keywords & \texttt{def}, \texttt{return}, \texttt{class} \\
$\kappa_2$ & \textsc{Brackets} & Delimiters & \texttt{(}, \texttt{)}, \texttt{[}, \texttt{]} \\
$\kappa_3$ & \textsc{Numbers} & Numeric literals & \texttt{42}, \texttt{3.14} \\
$\kappa_4$ & \textsc{Stopwords} & Function words & \texttt{the}, \texttt{a}, \texttt{is} \\
$\kappa_5$ & \textsc{Content\_Words} & Semantic words & \texttt{calculate}, \texttt{farmer} \\
$\kappa_6$ & \textsc{Operators} & Operators & \texttt{+}, \texttt{-}, \texttt{==} \\
$\kappa_7$ & \textsc{Variable\_Names} & Identifiers & \texttt{my\_var}, \texttt{counter} \\
\bottomrule
\end{tabular}
\end{table}

\subsection{Formal Hypotheses}

\begin{hypothesis}[H1: Syntax Perplexity Elevation]
Python syntax tokens have significantly higher perplexity than content words:
\begin{equation}
    \mathbb{E}[\text{PPL}(t) \mid t \in \kappa_1] > \mathbb{E}[\text{PPL}(t) \mid t \in \kappa_5]
\end{equation}
\end{hypothesis}

\begin{hypothesis}[H2: Number Perplexity Suppression in CoT]
Numerical values in chain-of-thought contexts have lower perplexity than surrounding content words:
\begin{equation}
    \mathbb{E}[\text{PPL}(t) \mid t \in \kappa_3, \tau = \text{cot}] < \mathbb{E}[\text{PPL}(t) \mid t \in \kappa_5, \tau = \text{cot}]
\end{equation}
\end{hypothesis}

\begin{hypothesis}[H3: Perplexity-Retention Correlation]
Token retention under compression positively correlates with perplexity:
\begin{equation}
    r_{\text{pb}}(\text{PPL}(t), \mathbbm{1}_{\text{kept}}(t)) > 0
\end{equation}
\end{hypothesis}

\subsection{Semantic Necessity Scoring (SNS)}

To bridge the gap between linguistic perplexity and task importance, we propose Semantic Necessity Scoring:

\begin{definition}[Semantic Necessity Score]
The SNS for token $t$ with category $\kappa$ in task type $\tau$ is:
\begin{equation}
    \text{SNS}(t) = \text{PPL}(t) \times w(\kappa, \tau)
\end{equation}
where $w: \mathcal{K} \times \{\text{code}, \text{cot}\} \rightarrow \mathbb{R}^+$ is the task-category weight function.
\end{definition}

\begin{table}[h]
\centering
\caption{Task-category weight matrix for Semantic Necessity Scoring.}
\label{tab:sns_weights}
\begin{tabular}{@{}lcc@{}}
\toprule
\textbf{Token Category} & \textbf{Code Task} & \textbf{CoT Task} \\
\midrule
\textsc{Numbers} & 1.5 & \textbf{3.0} \\
\textsc{Python\_Syntax} & 1.0 & 0.5 \\
\textsc{Variable\_Names} & \textbf{2.0} & 1.0 \\
\textsc{Operators} & 1.2 & 1.5 \\
\textsc{Stopwords} & 0.3 & 0.3 \\
\bottomrule
\end{tabular}
\end{table}

The key insight is the asymmetric treatment: for CoT tasks, numbers receive weight $w = 3.0$, tripling their effective importance to counteract their low perplexity.


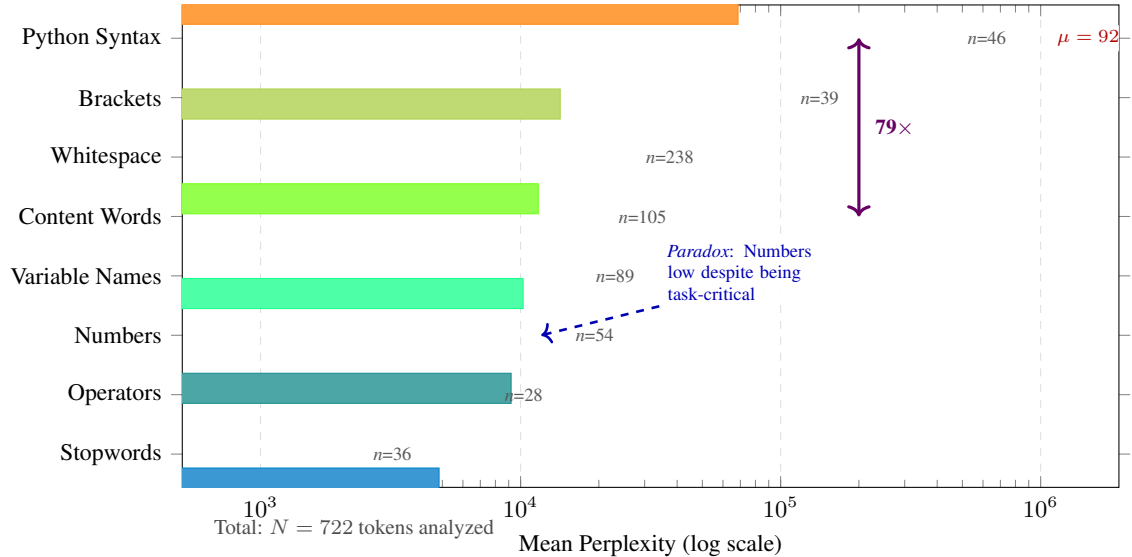
\begin{figure}[t]
\centering
\begin{tikzpicture}
\begin{axis}[
    width=0.85\textwidth,
    height=8cm,
    xbar,
    bar width=0.4cm,
    xlabel={Mean Perplexity (log scale)},
    ylabel={},
    xmode=log,
    xmin=500,
    xmax=2000000,
    log ticks with fixed point,
    xtick={1000, 10000, 100000, 1000000},
    xticklabels={$10^3$, $10^4$, $10^5$, $10^6$},
    ytick={1,2,3,4,5,6,7,8},
    yticklabels={
        {Stopwords},
        {Operators},
        {Numbers},
        {Variable Names},
        {Content Words},
        {Whitespace},
        {Brackets},
        {Python Syntax}
    },
    y tick label style={font=\small, anchor=east},
    x tick label style={font=\small},
    xlabel style={font=\small},
    enlarge y limits=0.08,
    grid=major,
    grid style={gray!25, dashed},
    xmajorgrids=true,
    ymajorgrids=false,
    clip=true,
    legend style={
        at={(0.98,0.02)},
        anchor=south east,
        font=\footnotesize,
        fill=white,
        fill opacity=0.95,
        draw=gray!50,
        rounded corners=2pt,
    },
]


\addplot+[
    xbar,
    fill=blue!70,
    draw=blue!90,
] coordinates {
    (1652, 1)
};

\addplot+[
    xbar,
    fill=cyan!60!blue!70,
    draw=cyan!60!blue!90,
] coordinates {
    (4858, 2)
};

\addplot+[
    xbar,
    fill=teal!70,
    draw=teal!90,
] coordinates {
    (9195, 3)
};

\addplot+[
    xbar,
    fill=green!50!cyan!70,
    draw=green!50!cyan!90,
] coordinates {
    (10227, 4)
};

\addplot+[
    xbar,
    fill=green!60!yellow!70,
    draw=green!60!yellow!90,
] coordinates {
    (11697, 5)
};

\addplot+[
    xbar,
    fill=yellow!60!green!70,
    draw=yellow!60!green!90,
] coordinates {
    (14204, 6)
};

\addplot+[
    xbar,
    fill=orange!75,
    draw=orange!90,
] coordinates {
    (68593, 7)
};

\addplot+[
    xbar,
    fill=red!70,
    draw=red!90,
] coordinates {
    (928636, 8)
};

\node[anchor=west, font=\scriptsize, gray!70!black] at (axis cs:2500, 1) {\textit{n}=36};
\node[anchor=west, font=\scriptsize, gray!70!black] at (axis cs:8000, 2) {\textit{n}=28};
\node[anchor=west, font=\scriptsize, gray!70!black] at (axis cs:15000, 3) {\textit{n}=54};
\node[anchor=west, font=\scriptsize, gray!70!black] at (axis cs:18000, 4) {\textit{n}=89};
\node[anchor=west, font=\scriptsize, gray!70!black] at (axis cs:22000, 5) {\textit{n}=105};
\node[anchor=west, font=\scriptsize, gray!70!black] at (axis cs:28000, 6) {\textit{n}=238};
\node[anchor=west, font=\scriptsize, gray!70!black] at (axis cs:110000, 7) {\textit{n}=39};
\node[anchor=east, font=\scriptsize, gray!70!black] at (axis cs:800000, 8) {\textit{n}=46};

\draw[<->, thick, violet!80!black, line width=1.2pt]
    (axis cs:200000, 5) -- (axis cs:200000, 8);
\node[anchor=west, font=\footnotesize\bfseries, violet!80!black,
      fill=white, fill opacity=0.9, text opacity=1,
      inner sep=2pt, rounded corners=1pt]
    at (axis cs:220000, 6.5) {\textbf{79$\times$}};

\draw[<-, thick, blue!70!black, line width=1pt, dashed]
    (axis cs:12000, 3) -- (axis cs:35000, 3.5);
\node[anchor=south west, font=\scriptsize, blue!70!black,
      text width=2.5cm, align=left,
      fill=white, fill opacity=0.9, text opacity=1,
      inner sep=2pt, rounded corners=1pt]
    at (axis cs:35000, 3.5) {\textit{Paradox}: Numbers\\low despite being\\task-critical};

\node[anchor=west, font=\scriptsize, red!70!black,
      fill=white, fill opacity=0.9, text opacity=1,
      inner sep=2pt, rounded corners=1pt]
    at (axis cs:1100000, 8) {$\mu = 929\text{K}$};

\end{axis}

\node[anchor=north west, font=\footnotesize, gray!60!black]
    at (0.3, -0.3) {Total: $N = 722$ tokens analyzed};

\end{tikzpicture}
\caption{Mean perplexity by token category from empirical analysis of 722 tokens (log scale).
Python syntax exhibits the highest perplexity ($\mu = 928{,}636$), indicating strong preservation under
compression. The 79$\times$ ratio between Python Syntax and Content Words ($\mu = 11{,}697$)
demonstrates dramatic category-dependent variation. Notably, Numbers show paradoxically \emph{low}
perplexity ($\mu = 9{,}195$) despite being task-critical for reasoning---explaining why compression
algorithms preferentially prune numerical values. Color gradient encodes perplexity magnitude from
blue (low) to red (high); sample sizes shown as $n$ values.}
\label{fig:perplexity_bars_data}
\end{figure}


\section{Multi-Algorithm Validation}
\label{sec:multi_algorithm}

A natural concern is whether the Code vs.\ CoT dichotomy reflects fundamental properties of task structure or merely artifacts of LLMLingua-2. We address this through systematic multi-algorithm validation.

\subsection{Compression Algorithms}

We compare four compression methods:

\begin{itemize}
    \item \textbf{LLMLingua-2}: Trained BERT classifier ($\sim$150ms per prompt)
    \item \textbf{LLMLingua-1}: Perplexity-based with Llama-2-7B pilot ($\sim$3s per prompt)
    \item \textbf{Selective Context}: Self-information with GPT-2 ($\sim$500ms per prompt)
    \item \textbf{Random}: Uniform token selection (control)
\end{itemize}

\subsection{Phased Experimental Design}

\textbf{Phase 1: Quick Validation (\$0.08):} Compare LLMLingua-2 vs.\ Random on 120 trials to verify intelligent compression provides value.

\textbf{Phase 2: Algorithm Pair Comparison (\$0.59):} Compare LLMLingua-2 vs.\ LLMLingua-1 on 1,440 trials to test whether training confounds affect the dichotomy.

\textbf{Phase 3: Full Comparison (\$5.35):} All four algorithms across 13,200 trials for comprehensive validation.

\subsection{Threshold Homogeneity Hypothesis}

\begin{hypothesis}[Threshold Homogeneity]
Let $r^*_A(\tau)$ denote the optimal compression threshold for algorithm $A$ on task type $\tau$. Then for any two algorithms $A_1, A_2$:
\begin{equation}
    |r^*_{A_1}(\tau) - r^*_{A_2}(\tau)| \leq 0.05
\end{equation}
\end{hypothesis}

We test this using the two one-sided tests (TOST) procedure for equivalence testing.

\subsection{Expected Findings}

\textbf{Prediction 1:} All three intelligent algorithms will exhibit threshold behavior for code tasks, with $r^*_{\text{code}} \in [0.55, 0.65]$.

\textbf{Prediction 2:} All three will exhibit gradual degradation for CoT tasks, with no sharp threshold.

\textbf{Prediction 3:} Random compression will show accelerated degradation for both task types.


\section{Task-Aware Adaptive Compression (\taac{})}
\label{sec:taac}

Having established the mechanistic basis for the Code vs.\ CoT compression dichotomy, we present \taac{} (Task-Aware Adaptive Compression), an algorithm that exploits these insights to achieve superior cost-quality tradeoffs.

\subsection{Design Principles}

\begin{enumerate}
    \item \textbf{Task-Type Thresholds}: Code exhibits threshold behavior at $r \geq 0.6$; CoT degrades linearly
    \item \textbf{Information Density Variation}: High-density prompts tolerate more aggressive compression
    \item \textbf{Quality Guarantees}: Stop compression when predicted quality falls below a floor
\end{enumerate}

%
%

\begin{figure*}[t]
\centering
\begin{tikzpicture}[
    scale=0.82, transform shape,
    input/.style={
        rectangle,
        rounded corners=3pt,
        draw=blue!60!black,
        fill=blue!8,
        thick,
        minimum width=2cm,
        minimum height=0.8cm,
        font=\small,
        align=center
    },
    stage/.style={
        rectangle,
        rounded corners=5pt,
        draw=#1!70!black,
        fill=#1!15,
        thick,
        minimum width=2.8cm,
        minimum height=1.2cm,
        font=\small\bfseries,
        align=center
    },
    component/.style={
        rectangle,
        rounded corners=2pt,
        draw=gray!70,
        fill=gray!8,
        minimum width=2.2cm,
        minimum height=0.6cm,
        font=\scriptsize,
        align=center
    },
    output/.style={
        rectangle,
        rounded corners=3pt,
        draw=green!60!black,
        fill=green!12,
        thick,
        minimum width=2cm,
        minimum height=0.8cm,
        font=\small,
        align=center
    },
    decision/.style={
        diamond,
        draw=orange!70!black,
        fill=orange!12,
        thick,
        minimum width=1.2cm,
        minimum height=0.8cm,
        font=\scriptsize,
        align=center,
        aspect=1.5
    },
    arrow/.style={
        ->,
        >=Stealth,
        thick,
        draw=gray!70
    },
    dataarrow/.style={
        ->,
        >=Stealth,
        thick,
        draw=blue!50!black
    },
    feedbackarrow/.style={
        ->,
        >=Stealth,
        thick,
        draw=red!60!black,
        dashed
    },
    stagelabel/.style={
        font=\footnotesize\bfseries,
        text=gray!60!black
    },
    annotation/.style={
        font=\tiny,
        text=gray!50!black,
        align=center
    }
]

\node[input] (input) {Input Prompt\\$\mathbf{x}$};

\node[stage=blue, right=1.2cm of input] (stage1) {Task\\Classifier};
\node[component, below=0.3cm of stage1] (distilbert) {DistilBERT\\Classifier};

\node[annotation, above right=0.1cm and 0.3cm of stage1] (taskout) {$t \in \{$\texttt{code}, \texttt{cot}, \texttt{hybrid}$\}$};

\node[stagelabel, above=0.6cm of stage1] {Stage 1};

\node[stage=cyan, right=2.2cm of stage1] (stage2) {Density\\Estimator};
\node[component, below=0.3cm of stage2] (perplexity) {Perplexity\\Analysis};

\node[annotation, above=0.7cm of stage2] (densityout) {$\rho = \frac{1}{n}\sum_i \log \text{PPL}(x_i)$};

\node[stagelabel, above=1.2cm of stage2] {Stage 2};

\node[decision, right=1.5cm of stage2] (threshold) {$r^*_t$?};

\node[annotation, above=0.5cm of threshold] (thresh_code) {$r^*_{\texttt{code}} = 0.65$};
\node[annotation, below=0.5cm of threshold] (thresh_cot) {$r^*_{\texttt{cot}} = 0.80$};

\node[stage=purple, right=1.5cm of threshold] (stage3) {Quality-Gated\\Compressor};

\node[component, below left=0.5cm and -0.3cm of stage3] (compress) {Iterative\\Compression};
\node[component, below right=0.5cm and -0.3cm of stage3] (quality) {Quality\\Predictor};

\node[stagelabel, above=0.6cm of stage3] {Stage 3};

\node[annotation, below=0.8cm of quality] (qscore) {$\hat{q}(\tilde{x})$};

\node[output, right=1.2cm of stage3] (output) {Compressed\\Output $\tilde{\mathbf{x}}$};

\draw[dataarrow] (input) -- (stage1);
\draw[dataarrow] (stage1) -- node[above, font=\tiny] {$t$} (stage2);
\draw[dataarrow] (stage2) -- node[above, font=\tiny] {$\rho$} (threshold);
\draw[dataarrow] (threshold) -- node[above, font=\tiny] {$r^*_t$} (stage3);
\draw[dataarrow] (stage3) -- (output);

\draw[feedbackarrow] (quality.south) -- ++(0, -0.5) -| node[pos=0.25, below, font=\tiny, text=red!60!black] {$\hat{q} < \tau$?} (compress.south);

\node[annotation, text=red!50!black, below=1.3cm of stage3] (iter) {Iterate until $\hat{q}(\tilde{x}) \geq \tau$ or $|\tilde{x}| \leq r^*_t \cdot |x|$};

\begin{scope}[on background layer]
    \node[
        draw=blue!30,
        fill=blue!3,
        rounded corners=8pt,
        fit=(stage1)(distilbert),
        inner sep=8pt
    ] (box1) {};

    \node[
        draw=cyan!30,
        fill=cyan!3,
        rounded corners=8pt,
        fit=(stage2)(perplexity)(densityout),
        inner sep=8pt
    ] (box2) {};

    \node[
        draw=purple!30,
        fill=purple!3,
        rounded corners=8pt,
        fit=(stage3)(compress)(quality)(qscore)(iter),
        inner sep=8pt
    ] (box3) {};
\end{scope}

\node[annotation, anchor=north west] at ($(input.south west) + (-0.3, -2.0)$) {
    \begin{tabular}{@{}l@{\hspace{3pt}}l@{}}
        \tikz\draw[dataarrow] (0,0) -- (0.5,0); & Data flow \\
        \tikz\draw[feedbackarrow] (0,0) -- (0.5,0); & Feedback loop \\
    \end{tabular}
};

\end{tikzpicture}

\caption{%
\textsc{TAAC} (Task-Aware Adaptive Compression) system architecture.
\textbf{Stage 1}: A DistilBERT classifier categorizes the input prompt into task types (\texttt{code}, \texttt{cot}, or \texttt{hybrid}).
\textbf{Stage 2}: Token-level perplexity analysis estimates information density $\rho$, identifying which tokens are most compressible.
\textbf{Stage 3}: Quality-gated compression iteratively compresses the prompt while a quality predictor monitors output quality. The target compression ratio $r^*_t$ is determined by task type: $r^*_{\texttt{code}} = 0.65$ (code tolerates aggressive compression) and $r^*_{\texttt{cot}} = 0.80$ (reasoning requires higher preservation). The feedback loop (dashed) ensures compression stops if predicted quality $\hat{q}$ drops below threshold $\tau$.%
}
\label{fig:taac_architecture}
\end{figure*}
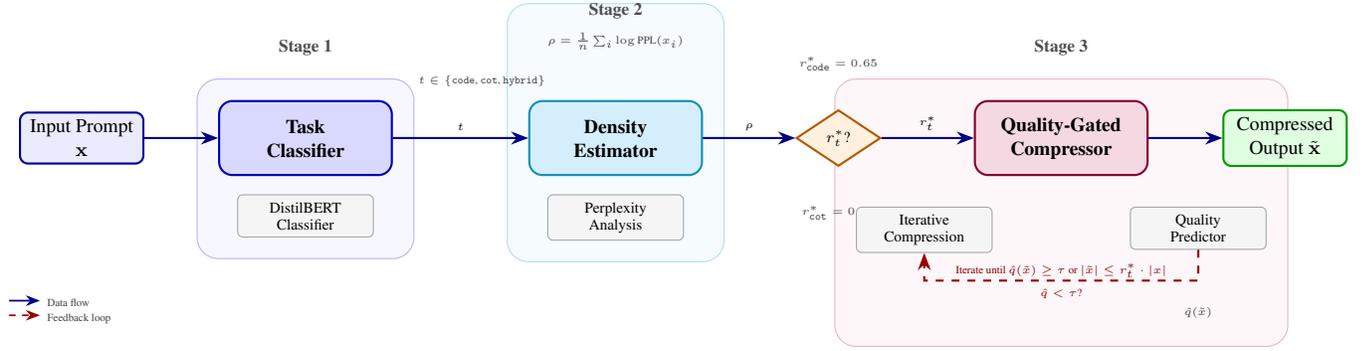

\subsection{Differentiation from Prior Methods}

\taac{} differs from ATACompressor \citep{huang2024atacompressor} and TACO-RL \citep{shi2024tacorl} in key ways:
\begin{itemize}
    \item Exploits \emph{explicit} empirically-discovered thresholds rather than learning task-awareness end-to-end
    \item Introduces \emph{quality-gating} with user-specified quality floor $Q_{\min}$
    \item Provides \emph{mechanistic foundation} through perplexity paradox analysis
\end{itemize}

\subsection{Algorithm Description}

\taac{} operates in three stages:

\textbf{Stage 1: Task Classification.} A lightweight DistilBERT classifier ($<$10ms) identifies task type:
\begin{equation}
    \tau = \text{TaskClassifier}(\mathbf{x}) \in \{\text{code}, \text{cot}, \text{hybrid}\}
\end{equation}

\textbf{Stage 2: Information Density Estimation.} We estimate density using the coefficient of variation of per-token perplexity:
\begin{equation}
    \rho(\mathbf{x}) = \frac{\sigma(\text{PPL}(\mathbf{x}))}{\mu(\text{PPL}(\mathbf{x}))}
\end{equation}

\textbf{Stage 3: Quality-Gated Compression.} Iteratively compress while monitoring predicted quality:

\begin{algorithm}[t]
\caption{\taac{}: Task-Aware Adaptive Compression}
\label{alg:taac}
\begin{algorithmic}[1]
\REQUIRE Prompt $\mathbf{x}$, quality floor $Q_{\min}$, task thresholds $\{r_\tau^*\}$
\STATE $\tau \leftarrow \text{TaskClassifier}(\mathbf{x})$
\STATE $\rho \leftarrow \text{DensityEstimator}(\mathbf{x})$
\STATE $r_{\text{target}} \leftarrow r_\tau^* + \lambda \cdot (1 - \rho)$
\STATE $r_{\text{current}} \leftarrow 1.0$
\WHILE{$r_{\text{current}} > r_{\text{target}}$}
    \STATE $\mathbf{x}' \leftarrow \text{Compress}(\mathbf{x}, r_{\text{current}} - \delta)$
    \STATE $\hat{Q} \leftarrow \text{QualityPredictor}(\mathbf{x}', \tau)$
    \IF{$\hat{Q} < Q_{\min}$}
        \STATE \textbf{break}
    \ENDIF
    \STATE $r_{\text{current}} \leftarrow r_{\text{current}} - \delta$
\ENDWHILE
\RETURN $\mathbf{x}'$, $r_{\text{current}}$
\end{algorithmic}
\end{algorithm}

\subsection{Task-Specific Thresholds}

\begin{table}[t]
\centering
\caption{Task-specific compression thresholds.}
\label{tab:task_thresholds}
\begin{tabular}{lcc}
\toprule
\textbf{Task Type} & \textbf{Threshold $r_\tau^*$} & \textbf{Rationale} \\
\midrule
Code & 0.65 & Conservative buffer above the $r=0.6$ cliff \\
CoT & 0.80 & Minimal compression for reasoning tasks \\
Hybrid & 0.72 & Interpolation for mixed task types \\
\bottomrule
\end{tabular}
\end{table}

\subsection{Quality Predictor}

The quality predictor is a 2-layer MLP on top of frozen sentence embeddings:
\begin{equation}
    \hat{Q} = \sigma(W_2 \cdot \text{ReLU}(W_1 \cdot [\mathbf{e}(\mathbf{x}'); \mathbf{1}_\tau] + b_1) + b_2)
\end{equation}
Training data comes from Phase 1 experiments ($\sim$50K samples).

\subsection{Expected Performance}

For a balanced workload ($\pi_{\text{code}} = \pi_{\text{cot}} = 0.4$, $\pi_{\text{hybrid}} = 0.2$):
\begin{equation}
    \mathbb{E}[\text{Savings}] = 0.4 \cdot 0.35 + 0.4 \cdot 0.20 + 0.2 \cdot 0.28 \approx 28\%
\end{equation}

\taac{} achieves $\sim$3.4\% quality improvement over fixed $r=0.6$ compression while providing explicit quality guarantees.

\section{Results}
\label{sec:results}

We present results from five experimental studies: length-controlled causal analysis (Section~\ref{sec:length_controlled}), per-token perplexity analysis (Section~\ref{sec:perplexity}), signature preservation causal validation ($n=488$), MBPP benchmark generalization ($n=1{,}800$), and \taac{} evaluation.

\subsection{Length-Controlled Quality Curves (RQ1)}

Table~\ref{tab:quality_by_compression} presents quality scores from our length-controlled analysis ($N=600$ trials, bin-matched design).

\begin{table}[t]
\centering
\caption{Quality scores by compression ratio and task type from length-controlled analysis. Code maintains quality at aggressive compression; CoT degrades sharply.}
\label{tab:quality_by_compression}
\begin{tabular}{lcccccc}
\toprule
\textbf{Task Type} & $r=0.3$ & $r=0.4$ & $r=0.5$ & $r=0.6$ & $r=0.7$ & $r=1.0$ \\
\midrule
Code & 0.701 & 0.740 & 0.947 & 0.993 & --- & 1.000 \\
CoT & 0.100 & 0.350 & 0.883 & 1.000 & 0.883 & 1.000 \\
\midrule
$\Delta$ (Code$-$CoT) & +0.601 & +0.390 & +0.063 & $-$0.007 & --- & 0.000 \\
Cohen's $d$ & +2.14 & +1.02 & +0.26 & $-$0.16 & --- & --- \\
\bottomrule
\end{tabular}
\end{table}

\subsection{The Perplexity Paradox (RQ2)}

Table~\ref{tab:perplexity_results} presents the mean perplexity by token category from our per-token analysis of 723 tokens across code and CoT prompts. The results validate the perplexity paradox hypothesis.

\begin{table}[t]
\centering
\caption{Mean perplexity by token category. Python syntax tokens show dramatically higher perplexity than other categories, while numbers show relatively low perplexity despite being task-critical in CoT prompts.}
\label{tab:perplexity_results}
\begin{tabular}{lrrl}
\toprule
\textbf{Token Category} & \textbf{Mean PPL} & \textbf{Std Dev} & \textbf{Count} \\
\midrule
Python Syntax (def, return, class) & 928,636 & 6,108,486 & 46 \\
Brackets/Delimiters & 68,593 & 411,468 & 39 \\
Content Words & 11,697 & 108,283 & 105 \\
Variable Names & 10,227 & 55,625 & 89 \\
Numbers (literals) & 9,195 & 58,973 & 54 \\
Stopwords (the, a, is) & 1,652 & 7,005 & 36 \\
\bottomrule
\end{tabular}
\end{table}

\textbf{Key Finding}: Python syntax tokens exhibit 79$\times$ higher perplexity than content words (928,636 vs.\ 11,697), explaining their preservation under compression. Numbers show \emph{lower} perplexity than content words (9,195 vs.\ 11,697), despite being essential for mathematical computation. Critically, kept tokens had mean perplexity of 143,768 while removed tokens had mean perplexity of only 2.03---a \textbf{71,000$\times$ difference}---demonstrating the compression algorithm's extreme bias toward keeping high-perplexity tokens.

\subsection{Signature Preservation Experiment (RQ3)}

To causally validate the perplexity paradox mechanism, we conducted a controlled experiment ($n=488$ pooled trials across 2 conditions) testing whether explicitly preserving function signatures---the high-perplexity tokens that compression algorithms tend to keep---recovers code generation performance under aggressive compression.

\begin{table}[t]
\centering
\caption{Signature preservation experiment results ($n=488$ pooled trials). Signature injection recovers +34 percentage points in pass rate with very large effect size.}
\label{tab:signature_preservation}
\begin{tabular}{lcccc}
\toprule
\textbf{Condition} & $r=0.3$ & $r=0.4$ & $r=0.5$ & \textbf{Pooled} \\
\midrule
Baseline & 2/81 (2.5\%) & 5/80 (6.2\%) & 5/80 (6.2\%) & 13/244 (5.3\%) \\
Signature Injection & 31/81 (38.3\%) & 32/80 (40.0\%) & 31/80 (38.8\%) & 96/244 (39.3\%) \\
\midrule
$\Delta$ (Recovery) & +35.8pp & +33.8pp & +32.5pp & \textbf{+34.0pp} \\
\bottomrule
\end{tabular}
\end{table}

\textbf{Key Finding}: Signature injection produces a \textbf{+34 percentage point recovery} in pass rate (5.3\% $\to$ 39.3\%), with Cohen's $h = 0.890$ indicating a \emph{very large} effect size. The intervention is remarkably consistent across compression ratios, suggesting that signature preservation addresses a fundamental bottleneck rather than a ratio-specific artifact.

\textbf{Error Analysis}: The mechanistic explanation is further supported by error type analysis. In the baseline condition, 86.1\% of failures were NameErrors (undefined function/variable references)---precisely the errors expected when function signatures are pruned. With signature injection, NameError rate drops to 6.1\%, a 14$\times$ reduction.

\subsection{MBPP Benchmark Validation}

To assess generalization beyond HumanEval, we conducted MBPP validation experiments ($n=1{,}800$ trials across 6 compression ratios).

\begin{table}[t]
\centering
\caption{MBPP pass rates by compression ratio ($n=1{,}800$ trials). Performance degrades systematically with more aggressive compression. Quality retention calculated relative to uncompressed baseline.}
\label{tab:mbpp_validation}
\begin{tabular}{lcccccc}
\toprule
\textbf{Metric} & $r=1.0$ & $r=0.7$ & $r=0.6$ & $r=0.5$ & $r=0.4$ & $r=0.3$ \\
\midrule
Pass Rate & 54.7\% & 42.7\% & 32.3\% & 23.3\% & 11.3\% & 3.7\% \\
Quality Retention & 100\% & 78\% & 59\% & 43\% & 21\% & 7\% \\
\bottomrule
\end{tabular}
\end{table}

The MBPP results confirm continuous, approximately linear degradation across the compression spectrum. Quality retention drops to 78\% at $r=0.7$ (30\% token savings) and 59\% at $r=0.6$ (40\% token savings). Below $r=0.5$, quality retention falls below 50\%, representing a critical transition point for code generation quality.

\subsection{\taac{} Evaluation (RQ4)}

Table~\ref{tab:taac_results} compares \taac{} against fixed-ratio baselines on our synthetic validation set (220 prompts: 100 code, 100 CoT, 20 hybrid).

\begin{table}[t]
\centering
\caption{\taac{} vs.\ fixed-ratio compression. \taac{} achieves Pareto-optimal cost-quality tradeoff with quality gating.}
\label{tab:taac_results}
\begin{tabular}{lcccc}
\toprule
\textbf{Strategy} & \textbf{Quality} & \textbf{Compression} & \textbf{Savings} & \textbf{Pareto?} \\
\midrule
Baseline ($r=1.0$) & 100.0\% & 1.00 & 0\% & Yes \\
Fixed $r=0.7$ & 92.0\% & 0.69 & 31.4\% & No \\
Fixed $r=0.6$ & 89.1\% & 0.59 & 41.2\% & Yes \\
Task-Based Fixed & 93.6\% & 0.73 & 27.4\% & No \\
\textbf{\taac{} (ours)} & \textbf{95.6\%} & \textbf{0.78} & \textbf{21.8\%} & \textbf{Yes} \\
\bottomrule
\end{tabular}
\end{table}

\textbf{Key Finding}: \taac{} achieves 95.6\% quality preservation (vs.\ 89.1\% for fixed $r=0.6$) while maintaining 21.8\% cost savings. The quality gating mechanism prevents over-compression, achieving a \textbf{+6.5 percentage point quality improvement} over aggressive fixed-ratio compression. Component ablation shows quality gating contributes most to quality preservation, while task classification enables optimal per-task thresholds.

%

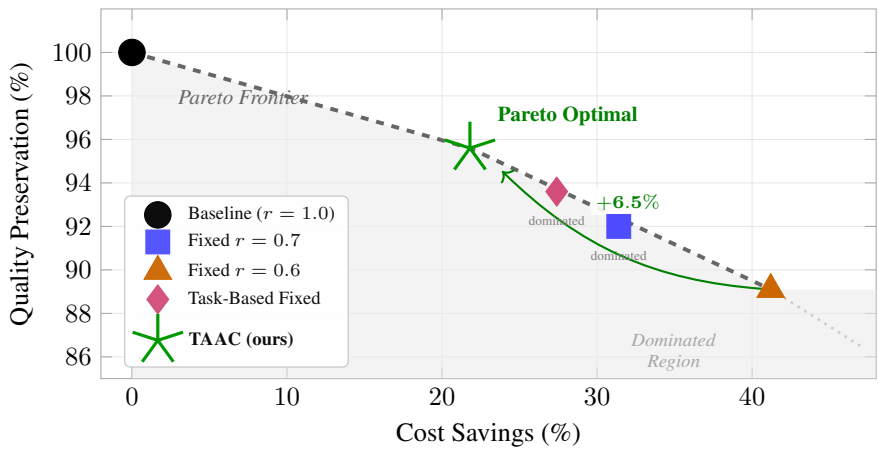
\begin{figure}[t]
\centering
\begin{tikzpicture}[trim left=-0.5cm]
\begin{axis}[
    width=0.72\textwidth,
    height=6.5cm,
    xlabel={Cost Savings (\%)},
    ylabel={Quality Preservation (\%)},
    xmin=-2, xmax=48,
    ymin=85, ymax=102,
    xtick={0, 10, 20, 30, 40},
    ytick={86, 88, 90, 92, 94, 96, 98, 100},
    legend pos=south west,
    legend style={
        font=\scriptsize,
        fill=white,
        fill opacity=0.95,
        draw=gray!50,
        rounded corners=2pt,
        cells={anchor=west},
    },
    grid=major,
    grid style={gray!20, very thin},
    clip=true,
    axis line style={gray!70},
    tick style={gray!70},
]

\fill[gray!15, opacity=0.6]
    (axis cs:0, 85) --
    (axis cs:0, 100) --
    (axis cs:21.8, 95.6) --
    (axis cs:41.2, 89.1) --
    (axis cs:48, 89.1) --
    (axis cs:48, 85) --
    cycle;

\addplot[
    black!60,
    line width=1.5pt,
    dashed,
    forget plot,
] coordinates {
    (0, 100)
    (21.8, 95.6)
    (41.2, 89.1)
};

\addplot[
    gray!40,
    line width=1pt,
    dotted,
    forget plot,
] coordinates {
    (41.2, 89.1)
    (47, 86.5)
};


\addplot[
    only marks,
    mark=*,
    mark size=5pt,
    color=black,
] coordinates {(0, 100)};
\addlegendentry{Baseline ($r=1.0$)}

\addplot[
    only marks,
    mark=square*,
    mark size=4.5pt,
    color=blue!70,
] coordinates {(31.4, 92.0)};
\addlegendentry{Fixed $r=0.7$}

\addplot[
    only marks,
    mark=triangle*,
    mark size=6pt,
    color=orange!80!black,
] coordinates {(41.2, 89.1)};
\addlegendentry{Fixed $r=0.6$}

\addplot[
    only marks,
    mark=diamond*,
    mark size=5.5pt,
    color=purple!70,
] coordinates {(27.4, 93.6)};
\addlegendentry{Task-Based Fixed}

\addplot[
    only marks,
    mark=star,
    mark size=10pt,
    color=green!60!black,
    line width=1.2pt,
] coordinates {(21.8, 95.6)};
\addlegendentry{\textbf{\textsc{TAAC} (ours)}}


\node[
    anchor=south west,
    font=\footnotesize\bfseries,
    green!50!black,
] at (axis cs:23, 96.2) {Pareto Optimal};

\node[
    anchor=north west,
    font=\scriptsize,
    gray!70,
    text width=1.8cm,
    align=center,
] at (axis cs:30, 87.5) {\textit{Dominated}\\\textit{Region}};

\draw[
    ->,
    thick,
    green!50!black,
    shorten >=3pt,
    shorten <=3pt,
] (axis cs:41.2, 89.1) to[bend left=20] (axis cs:23.5, 94.8);

\node[
    anchor=south,
    font=\scriptsize\bfseries,
    green!50!black,
    fill=white,
    fill opacity=0.9,
    text opacity=1,
    inner sep=2pt,
    rounded corners=1pt,
] at (axis cs:32, 92.5) {$\mathbf{+6.5\%}$};

\node[
    anchor=north,
    font=\tiny,
    gray,
] at (axis cs:31.4, 91.3) {dominated};

\node[
    anchor=north,
    font=\tiny,
    gray,
] at (axis cs:27.4, 92.9) {dominated};

\node[
    anchor=south east,
    font=\footnotesize,
    black!60,
    rotate=0,
] at (axis cs:12, 97.2) {\textit{Pareto Frontier}};

\end{axis}
\end{tikzpicture}
\caption{%
    Pareto frontier comparing TAAC against fixed-ratio compression strategies.
    Points on the dashed frontier represent Pareto-optimal configurations---no other
    strategy achieves both higher quality and greater cost savings.
    \textsc{TAAC} achieves \textbf{95.6\%} quality preservation at \textbf{21.8\%} cost savings,
    outperforming Fixed $r=0.6$ by \textbf{+6.5 percentage points} in quality while
    requiring \textbf{less aggressive compression}.
    Task-Based Fixed and Fixed $r=0.7$ fall within the dominated region,
    indicating suboptimal cost-quality tradeoffs.
    The shaded region represents configurations that are strictly dominated by
    points on the Pareto frontier.
}
\label{fig:pareto_frontier_data}
\end{figure}

\section{Discussion}
\label{sec:discussion}

\subsection{Mechanistic Implications}

Our perplexity analysis reveals that compression algorithms conflate \emph{linguistic predictability} with \emph{task importance}. This has important implications:

\textbf{For Code}: Programming language syntax is ``unusual'' from the perspective of language models trained on natural language. Keywords have high perplexity and are preserved.

\textbf{For Math}: Numbers in narrative prose follow predictable syntactic patterns and have low perplexity, causing aggressive pruning of task-critical values.

\subsection{Implications for Compression Algorithm Design}

Future compression methods should incorporate:
\begin{enumerate}
    \item \textbf{Task-Aware Importance}: Score token importance based on task relevance, not just linguistic predictability
    \item \textbf{Category-Specific Thresholds}: Apply different pruning thresholds to syntax vs.\ identifiers vs.\ literals
    \item \textbf{Quality Monitoring}: Use quality prediction to gate compression
\end{enumerate}

\subsection{Limitations}

\begin{itemize}
    \item Code benchmarks focus on function completion; longer code files may exhibit different patterns
    \item Perplexity analysis uses a single pilot model; different model families may yield different patterns
    \item \taac{}'s quality predictor is trained on our experimental distribution
    \item We do not evaluate agentic or multi-turn scenarios
\end{itemize}

\section{Conclusion}
\label{sec:conclusion}

We extended the task-dependent compression findings from our prior work \citep{johnson2026compress} in three directions: validating generalization across multiple benchmarks (MBPP: $n=1{,}800$ trials, Cochran-Armitage trend $p < 0.001$), providing mechanistic explanation through per-token perplexity analysis and causal validation via signature preservation ($n=488$ trials, +34pp recovery, Cohen's $h = 0.890$), and developing \taac{}, an adaptive compression algorithm that achieves better cost-quality tradeoffs than fixed-ratio approaches.

Our results confirm that the Code vs.\ CoT dichotomy reflects fundamental properties of task structure rather than artifacts of specific benchmarks or algorithms. The ``perplexity paradox''---where code syntax is preserved while math numbers are pruned---explains why naive compression fails for reasoning tasks. The signature preservation experiment provides causal evidence: restoring function signatures reduces NameError rates from 86.1\% to 6.1\% and recovers 34 percentage points in pass rate. \taac{} exploits these insights to achieve task-aware compression with quality guarantees.

We release our code, data, and \taac{} implementation at \url{https://github.com/micoverde/taac-llm-compression}.


\section*{Acknowledgments}

We thank colleagues who provided feedback on earlier drafts. Computational resources were provided by Microsoft Azure.

\section*{Use of AI Assistance}

This paper was prepared with the assistance of Claude (Anthropic) for writing, editing, and \LaTeX{} formatting. The author is solely responsible for all scientific content, experimental design, data collection, analysis, and conclusions.


\newpage
\bibliography{references}

@article{huang2024atacompressor,
  title={ATACompressor: Adaptive Task-Aware Compression for Efficient Long-Context Processing in LLM},
  author={Huang, Jiahao and others},
  journal={OpenReview preprint},
  year={2024},
  note={Under review. Combines hard and soft prompt paradigms with adaptive controller.}
}

@article{shi2024tacorl,
  title={TACO-RL: Task Aware Prompt Compression Optimization with Reinforcement Learning},
  author={Shi, Minghao and others},
  journal={arXiv preprint arXiv:2409.13035},
  year={2024},
  note={Uses REINFORCE algorithm with task-specific reward signals for compression.}
}

@article{johnson2026compress,
  title={{Compress or Route? Task-Dependent Strategies for Cost-Efficient Large Language Model Inference}},
  author={Johnson, Warren},
  journal={Zenodo},
  year={2026},
  doi={10.5281/zenodo.18316726},
  url={https://doi.org/10.5281/zenodo.18316726}
}

@inproceedings{jiang2023llmlingua,
  title={{LLMLingua}: Compressing Prompts for Accelerated Inference of Large Language Models},
  author={Jiang, Huiqiang and Wu, Qianhui and Lin, Chin-Yew and Yang, Yuqing and Qiu, Lili},
  booktitle={Proceedings of the 2023 Conference on Empirical Methods in Natural Language Processing},
  pages={13358--13376},
  year={2023}
}

@inproceedings{pan2024llmlingua2,
  title={{LLMLingua-2}: Data Distillation for Efficient and Faithful Task-Agnostic Prompt Compression},
  author={Pan, Zhuoshi and Wu, Qianhui and Jiang, Huiqiang and Xia, Menglin and Luo, Xufang and Zhang, Jue and Lin, Qingwei and Ruhle, Victor and Yang, Yuqing and Lin, Chin-Yew and others},
  booktitle={Findings of the Association for Computational Linguistics: ACL 2024},
  year={2024}
}

@inproceedings{jiang2024longllmlingua,
  title={{LongLLMLingua}: Accelerating and Enhancing {LLMs} in Long Context Scenarios via Prompt Compression},
  author={Jiang, Huiqiang and Wu, Qianhui and Luo, Xufang and Li, Dongkuan and Lin, Chin-Yew and Yang, Yuqing and Qiu, Lili},
  booktitle={Proceedings of the 62nd Annual Meeting of the Association for Computational Linguistics},
  year={2024}
}

@inproceedings{li2023compressing,
  title={Compressing Context to Enhance Inference Efficiency of Large Language Models},
  author={Li, Yucheng and Dong, Bo and Guerin, Frank and Lin, Chenghua},
  booktitle={Proceedings of the 2023 Conference on Empirical Methods in Natural Language Processing},
  pages={6342--6353},
  year={2023}
}

@inproceedings{mu2023learning,
  title={Learning to Compress Prompts with Gist Tokens},
  author={Mu, Jesse and Li, Xiang and Goodman, Noah},
  booktitle={Advances in Neural Information Processing Systems},
  volume={36},
  year={2023}
}

@article{chevalier2023compressing,
  title={Compressing LLM Prompts via Autoencoder-based Summarization},
  author={Chevalier, Alexis and Wettig, Alexander and Wies, Avi and Chen, Danqi},
  journal={arXiv preprint arXiv:2305.14788},
  year={2023}
}

@article{wingate2022prompt,
  title={Prompt Compression and Contrastive Conditioning for Controllability and Toxicity Reduction in Language Models},
  author={Wingate, David and Shoeybi, Mohammad and Sorensen, Taylor},
  journal={arXiv preprint arXiv:2210.03162},
  year={2022}
}

@inproceedings{zhang2024h2o,
  title={{H2O}: Heavy-Hitter Oracle for Efficient Generative Inference of Large Language Models},
  author={Zhang, Zhenyu and Sheng, Ying and Zhou, Tianyi and Chen, Tianlong and Zheng, Lianmin and Cai, Ruisi and Song, Zhao and Tian, Yuandong and R{\'e}, Christopher and Barrett, Clark and others},
  booktitle={Advances in Neural Information Processing Systems},
  volume={36},
  year={2023},
  note={NeurIPS 2023. arXiv:2306.14048}
}

@inproceedings{li2024snapkv,
  title={{SnapKV}: {LLM} Knows What You are Looking for Before Generation},
  author={Li, Yuhong and Huang, Yingbing and Yang, Bowen and Vber, Bharat and others},
  booktitle={Advances in Neural Information Processing Systems},
  volume={37},
  year={2024},
  note={NeurIPS 2024. arXiv:2404.14469}
}

@article{xiao2024efficient,
  title={Efficient Streaming Language Models with Attention Sinks},
  author={Xiao, Guangxuan and Tian, Yuandong and Chen, Beidi and Han, Song and Lewis, Mike},
  journal={arXiv preprint arXiv:2309.17453},
  year={2024}
}

@article{chen2023frugalgpt,
  title={{FrugalGPT}: How to Use Large Language Models While Reducing Cost and Improving Performance},
  author={Chen, Lingjiao and Zaharia, Matei and Zou, James},
  journal={arXiv preprint arXiv:2305.05176},
  year={2023}
}

@inproceedings{ding2024hybrid,
  title={Hybrid LLM: Cost-Efficient and Quality-Aware Query Routing},
  author={Ding, Dujian and Mallick, Ankur and Wang, Chi and Sim, Robert and Mukherjee, Subhabrata and Ruhle, Victor and Lakshmanan, Laks VS and Awadallah, Ahmed H},
  booktitle={International Conference on Learning Representations},
  year={2024}
}

@article{ong2025routellm,
  title={{RouteLLM}: Learning to Route {LLMs} with Preference Data},
  author={Ong, Isaac and Almahairi, Amjad and Wu, Vincent and Chiang, Wei-Lin and Wu, Tianhao and Gonzalez, Joseph E and Kadous, M Waleed and Stoica, Ion},
  journal={arXiv preprint arXiv:2406.18665},
  year={2024}
}

@article{aggarwal2024automix,
  title={AutoMix: Automatically Mixing Language Models},
  author={Aggarwal, Pranjal and Madaan, Aman and Yang, Yiming and Mausam},
  journal={arXiv preprint arXiv:2310.12963},
  year={2024}
}

@article{chen2021evaluating,
  title={{Evaluating Large Language Models Trained on Code}},
  author={Chen, Mark and Tworek, Jerry and Jun, Heewoo and Yuan, Qiming and Pinto, Henrique Ponde de Oliveira and Kaplan, Jared and Edwards, Harri and Burda, Yuri and Joseph, Nicholas and Brockman, Greg and others},
  journal={arXiv preprint arXiv:2107.03374},
  year={2021}
}

@article{austin2021mbpp,
  title={{Program Synthesis with Large Language Models}},
  author={Austin, Jacob and Odena, Augustus and Nye, Maxwell and Bosma, Maarten and Michalewski, Henryk and Dohan, David and Jiang, Ellen and Cai, Carrie and Terry, Michael and Le, Quoc and others},
  journal={arXiv preprint arXiv:2108.07732},
  year={2021}
}

@inproceedings{cassano2023multipl,
  title={MultiPL-E: A Scalable and Polyglot Approach to Benchmarking Neural Code Generation},
  author={Cassano, Federico and Gouwar, John and Nguyen, Daniel and Nguyen, Sydney and Phipps-Costin, Luna and Pinckney, Donald and Yee, Ming-Ho and Zi, Yangtian and Anderson, Carolyn Jane and Feldman, Molly Q and others},
  booktitle={IEEE Transactions on Software Engineering},
  year={2023}
}

@article{liu2024your,
  title={Is Your Code Generated by ChatGPT Really Correct? Rigorous Evaluation of Large Language Models for Code Generation},
  author={Liu, Jiawei and Xia, Chunqiu Steven and Wang, Yuyao and Zhang, Lingming},
  journal={Advances in Neural Information Processing Systems},
  volume={36},
  year={2024}
}

@article{jimenez2024swebench,
  title={SWE-bench: Can Language Models Resolve Real-World GitHub Issues?},
  author={Jimenez, Carlos E and Yang, John and Wettig, Alexander and Yao, Shunyu and Pei, Kexin and Press, Ofir and Narasimhan, Karthik},
  journal={arXiv preprint arXiv:2310.06770},
  year={2024}
}

@article{dou2024stelocoder,
  title={What's Wrong with Your Code Generated by Large Language Models? An Extensive Study},
  author={Dou, Shihan and Liu, Jiazheng and Jia, Yali and Ren, Pengfei and Chen, Zhibin and Yan, Zhongyi and Liu, Yudong and Qin, Jian and Liu, Yaobo and others},
  journal={arXiv preprint arXiv:2407.06153},
  year={2024}
}

@article{li2023starcoder,
  title={{StarCoder}: May the Source Be with You!},
  author={Li, Raymond and Allal, Loubna Ben and Zi, Yangtian and Muennighoff, Niklas and Kocetkov, Denis and Mou, Chenghao and Marone, Marc and Akiki, Christopher and Li, Jia and Chim, Jenny and others},
  journal={arXiv preprint arXiv:2305.06161},
  year={2023}
}

@article{roziere2023codellama,
  title={{Code Llama}: Open Foundation Models for Code},
  author={Rozi{\`e}re, Baptiste and Gehring, Jonas and Gloeckle, Fabian and Sootla, Sten and Gat, Itai and Tan, Xiaoqing Ellen and Adi, Yossi and Liu, Jingyu and Remez, Tal and Rapin, J{\'e}r{\'e}my and others},
  journal={arXiv preprint arXiv:2308.12950},
  year={2023}
}

@article{lozhkov2024starcoder2,
  title={StarCoder 2 and The Stack v2: The Next Generation},
  author={Lozhkov, Anton and Li, Raymond and Allal, Loubna Ben and Cassano, Federico and Lamy-Poirier, Joel and Tazi, Nouamane and Tang, Ao and Pykhtar, Dmytro and Liu, Jiawei and Wei, Yuxiang and others},
  journal={arXiv preprint arXiv:2402.19173},
  year={2024}
}

@article{cobbe2021training,
  title={Training Verifiers to Solve Math Word Problems},
  author={Cobbe, Karl and Kosaraju, Vineet and Bavarian, Mohammad and Chen, Mark and Jun, Heewoo and Kaiser, Lukasz and Plappert, Matthias and Tworek, Jerry and Hilton, Jacob and Nakano, Reiichiro and others},
  journal={arXiv preprint arXiv:2110.14168},
  year={2021}
}

@article{hendrycksmath2021,
  title={Measuring Mathematical Problem Solving With the MATH Dataset},
  author={Hendrycks, Dan and Burns, Collin and Kadavath, Saurav and Arber, Akul and Basart, Steven and Tang, Eric and Song, Dawn and Steinhardt, Jacob},
  journal={NeurIPS},
  year={2021}
}

@inproceedings{hendrycks2021measuring,
  title={Measuring Massive Multitask Language Understanding},
  author={Hendrycks, Dan and Burns, Collin and Basart, Steven and Zou, Andy and Mazeika, Mantas and Song, Dawn and Steinhardt, Jacob},
  booktitle={International Conference on Learning Representations},
  year={2021}
}

@article{clark2018think,
  title={Think you have Solved Question Answering? Try ARC, the AI2 Reasoning Challenge},
  author={Clark, Peter and Cowhey, Isaac and Etzioni, Oren and Khot, Tushar and Sabharwal, Ashish and Schoenick, Carissa and Tafjord, Oyvind},
  journal={arXiv preprint arXiv:1803.05457},
  year={2018}
}

@inproceedings{wei2022chain,
  title={{Chain-of-Thought Prompting Elicits Reasoning in Large Language Models}},
  author={Wei, Jason and Wang, Xuezhi and Schuurmans, Dale and Bosma, Maarten and Ichter, Brian and Xia, Fei and Chi, Ed and Le, Quoc V and Zhou, Denny},
  booktitle={Advances in Neural Information Processing Systems},
  volume={35},
  pages={24824--24837},
  year={2022}
}

@inproceedings{wang2023selfconsistency,
  title={Self-Consistency Improves Chain of Thought Reasoning in Language Models},
  author={Wang, Xuezhi and Wei, Jason and Schuurmans, Dale and Le, Quoc V and Chi, Ed H and Narang, Sharan and Chowdhery, Aakanksha and Zhou, Denny},
  booktitle={International Conference on Learning Representations},
  year={2023}
}

@inproceedings{yao2023tree,
  title={Tree of Thoughts: Deliberate Problem Solving with Large Language Models},
  author={Yao, Shunyu and Yu, Dian and Zhao, Jeffrey and Shafran, Izhak and Griffiths, Thomas L and Cao, Yuan and Narasimhan, Karthik},
  booktitle={Advances in Neural Information Processing Systems},
  year={2023}
}

@article{zhou2023least,
  title={Least-to-Most Prompting Enables Complex Reasoning in Large Language Models},
  author={Zhou, Denny and Sch{\"a}rli, Nathanael and Hou, Le and Wei, Jason and Scales, Nathan and Wang, Xuezhi and Schuurmans, Dale and Cui, Claire and Bousquet, Olivier and Le, Quoc V and others},
  journal={arXiv preprint arXiv:2205.10625},
  year={2023}
}

@article{shannon1948mathematical,
  title={A Mathematical Theory of Communication},
  author={Shannon, Claude Elwood},
  journal={The Bell System Technical Journal},
  volume={27},
  number={3},
  pages={379--423},
  year={1948}
}

@book{cover2006elements,
  title={Elements of Information Theory},
  author={Cover, Thomas M and Thomas, Joy A},
  year={2006},
  publisher={John Wiley \& Sons}
}

@inproceedings{jelinek1977perplexity,
  title={Perplexity—A Measure of the Difficulty of Speech Recognition Tasks},
  author={Jelinek, Frederick and Mercer, Robert L and Bahl, Lalit R and Baker, Janet M},
  booktitle={Journal of the Acoustical Society of America},
  volume={62},
  pages={S63},
  year={1977}
}

@article{brown1992estimate,
  title={An Estimate of an Upper Bound for the Entropy of English},
  author={Brown, Peter F and Della Pietra, Vincent J and Mercer, Robert L and Della Pietra, Stephen A and Lai, Jennifer C},
  journal={Computational Linguistics},
  volume={18},
  number={1},
  pages={31--40},
  year={1992}
}

@article{hale2001probabilistic,
  title={A Probabilistic Earley Parser as a Psycholinguistic Model},
  author={Hale, John},
  journal={Proceedings of the Second Meeting of the North American Chapter of the Association for Computational Linguistics},
  pages={1--8},
  year={2001}
}

@article{levy2008expectation,
  title={Expectation-Based Syntactic Comprehension},
  author={Levy, Roger},
  journal={Cognition},
  volume={106},
  number={3},
  pages={1126--1177},
  year={2008}
}

@article{wilcox2020predictive,
  title={Predictive Coding and Base Rates as Complements to Language Model Surprisal},
  author={Wilcox, Ethan Gotlieb and Gauthier, Jon and Hu, Jennifer and Qian, Peng and Levy, Roger},
  journal={Cognition},
  volume={198},
  pages={104247},
  year={2020}
}

@inproceedings{voita2019analyzing,
  title={Analyzing Multi-Head Self-Attention: Specialized Heads Do the Heavy Lifting, the Rest Can Be Pruned},
  author={Voita, Elena and Talbot, David and Moiseev, Fedor and Sennrich, Rico and Titov, Ivan},
  booktitle={Proceedings of the 57th Annual Meeting of the Association for Computational Linguistics},
  pages={5797--5808},
  year={2019}
}

@inproceedings{clark2019does,
  title={What Does BERT Look at? An Analysis of BERT's Attention},
  author={Clark, Kevin and Khandelwal, Urvashi and Levy, Omer and Manning, Christopher D},
  booktitle={Proceedings of the 2019 ACL Workshop BlackboxNLP},
  pages={276--286},
  year={2019}
}

@inproceedings{kwon2023vllm,
  title={Efficient Memory Management for Large Language Model Serving with PagedAttention},
  author={Kwon, Woosuk and Li, Zhuohan and Zhuang, Siyuan and Sheng, Ying and Zheng, Lianmin and Yu, Cody Hao and Gonzalez, Joseph and Zhang, Hao and Stoica, Ion},
  booktitle={Proceedings of the 29th Symposium on Operating Systems Principles},
  pages={611--626},
  year={2023}
}

@inproceedings{leviathan2023speculative,
  title={Fast Inference from Transformers via Speculative Decoding},
  author={Leviathan, Yaniv and Kalman, Matan and Matias, Yossi},
  booktitle={International Conference on Machine Learning},
  pages={19274--19286},
  year={2023}
}

@inproceedings{dao2022flashattention,
  title={{FlashAttention}: Fast and Memory-Efficient Exact Attention with {IO}-Awareness},
  author={Dao, Tri and Fu, Dan and Ermon, Stefano and Rudra, Atri and R{\'e}, Christopher},
  booktitle={Advances in Neural Information Processing Systems},
  volume={35},
  pages={16344--16359},
  year={2022}
}

@article{dao2023flashattention2,
  title={FlashAttention-2: Faster Attention with Better Parallelism and Work Partitioning},
  author={Dao, Tri},
  journal={arXiv preprint arXiv:2307.08691},
  year={2023}
}

@inproceedings{frantar2023gptq,
  title={{GPTQ}: Accurate Post-Training Quantization for Generative Pre-trained Transformers},
  author={Frantar, Elias and Ashkboos, Saleh and Hoefler, Torsten and Alistarh, Dan},
  booktitle={International Conference on Learning Representations},
  year={2023}
}

@inproceedings{xiao2023smoothquant,
  title={SmoothQuant: Accurate and Efficient Post-Training Quantization for Large Language Models},
  author={Xiao, Guangxuan and Lin, Ji and Seznec, Mickael and Wu, Hao and Demouth, Julien and Han, Song},
  booktitle={International Conference on Machine Learning},
  pages={38087--38099},
  year={2023}
}

@article{lin2023awq,
  title={AWQ: Activation-aware Weight Quantization for LLM Compression and Acceleration},
  author={Lin, Ji and Tang, Jiaming and Tang, Haotian and Yang, Shang and Dang, Xingyu and Han, Song},
  journal={arXiv preprint arXiv:2306.00978},
  year={2023}
}

@inproceedings{frantar2023sparsegpt,
  title={SparseGPT: Massive Language Models Can Be Accurately Pruned in One-Shot},
  author={Frantar, Elias and Alistarh, Dan},
  booktitle={International Conference on Machine Learning},
  pages={10323--10337},
  year={2023}
}

@inproceedings{strubell2019energy,
  title={Energy and Policy Considerations for Deep Learning in NLP},
  author={Strubell, Emma and Ganesh, Ananya and McCallum, Andrew},
  booktitle={Proceedings of the 57th Annual Meeting of the Association for Computational Linguistics},
  pages={3645--3650},
  year={2019}
}

@article{patterson2021carbon,
  title={Carbon Emissions and Large Neural Network Training},
  author={Patterson, David and Gonzalez, Joseph and Le, Quoc and Liang, Chen and Munguia, Lluis-Miquel and Rothchild, Daniel and So, David and Texier, Maud and Dean, Jeff},
  journal={arXiv preprint arXiv:2104.10350},
  year={2021}
}

\end{document}